\documentclass[11pt,a4paper]{article}
\usepackage[hyperref]{acl2020}
\usepackage{times}
\usepackage{latexsym}
\usepackage{graphicx}
\usepackage{amsmath, amssymb, mathtools, bm}
\usepackage{todonotes}
\usepackage{hyperref}
\usepackage{enumitem}
\usepackage{booktabs}
\usepackage{colortbl}

\usepackage{microtype}

\aclfinalcopy 

\title{Natural Language Inference with Mixed Effects}

\author{William Gantt\thanks{\; Equal contribution} \\
  University of Rochester \\\And
  Benjamin Kane\footnotemark[1] \\
  University of Rochester \\\And
  Aaron Steven White \\
  University of Rochester }

\date{}

\newcommand{\xb}{\mathbf{x}}
\newcommand{\zb}{\mathbf{z}}
\newcommand{\reals}{\mathbb{R}}

\newcommand{\annotators}{\mathcal{A}}
\newcommand{\thpair}[1]{\langle T_{#1}, H_{#1}\rangle}
\newcommand{\labels}{\mathcal{Y}}

\definecolor{light-gray}{gray}{0.97}
\definecolor{bvd-green}{HTML}{B7FEA6}
\definecolor{bvd-blue}{HTML}{A6D1FE}
\definecolor{bvd-gray}{HTML}{EEE9FD}
\definecolor{ao-eng}{rgb}{0.0, 0.5, 0.0}

\begin{document}
\maketitle
\begin{abstract}
There is growing evidence that the prevalence of disagreement in the raw annotations used to construct natural language inference datasets makes the common practice of aggregating those annotations to a single label problematic. We propose a generic method that allows one to skip the aggregation step and train on the raw annotations directly without subjecting the model to unwanted noise that can arise from annotator response biases. We demonstrate that this method, which generalizes the notion of a \textit{mixed effects model} by incorporating \textit{annotator random effects} into any existing neural model, improves performance over models that do not incorporate such effects.  
\end{abstract}

\section{Introduction}
\label{sec:introduction}

A common method for constructing natural language inference (NLI) datasets is (i) to generate text-hypothesis pairs using some method---commonly, crowd-sourced hypothesis elicitation given a text from some existing resource \citep{bowman-etal-2015-large, williams-etal-2018-broad} or automated text-hypothesis generation \citep{zhang-etal-2017-ordinal}; (ii) to collect crowd-sourced judgments about inference from the text to the hypothesis; and (iii) to aggregate the possibly multiple annotations provided for a single text-hypothesis pair into a single label. This final step follows common practice across annotation tasks in NLP; but for NLI in particular, there is growing evidence that it is problematic due to disagreement among annotators that is not captured by the probabilistic outputs of standard NLI models \citep[][]{pavlick-kwiatkowski-2019-inherent}.

\begin{figure}
    \centering
    \includegraphics[width=\columnwidth]{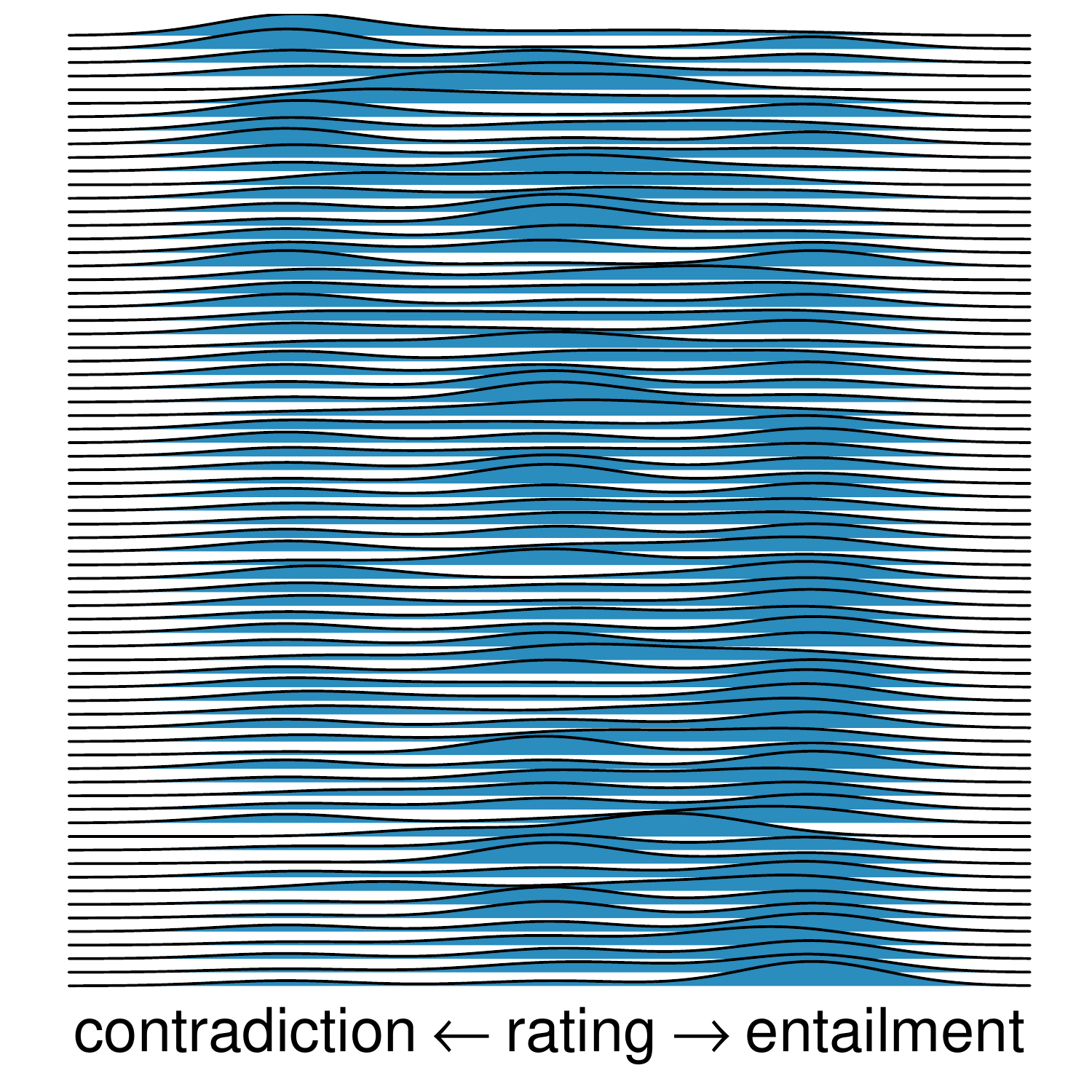}
    \vspace{-8mm}
    \caption{Distribution of [-50, 50] slider ratings (by annotator) for the same 20 NLI pairs in \citeauthor{pavlick-kwiatkowski-2019-inherent}'s dataset (\href{https://github.com/epavlick/NLI-variation-data/blob/master/sentence-pair-analysis/raw/batch1.csv}{batch 1}, described in their \S{}3).}
    \label{fig:variability-example}
    \vspace{-6mm}
\end{figure}

One way to capture this disagreement would be to directly model the variability in the raw annotations. But this approach presents a challenge: it can be difficult to assess how much disagreement arises from disagreement about the interpretation of a text-hypothesis pair and how much is due to biases that annotators bring to the task. Such biases can be extreme. For instance, \autoref{fig:variability-example} plots the distribution by annotator of [-50, 50] ratings---with -50 clear contradiction and 50 clear entailment---for the same 20 NLI pairs in \citeauthor{pavlick-kwiatkowski-2019-inherent}'s dataset. Despite describing responses to the same items, the distributions are quite variable, suggesting variability in how annotators approach the task. This difference in approach may be relatively shallow---e.g. given some true label (or distribution thereon), annotators merely differ in their mapping of that value to the response scale---or they may be quite deep---e.g. annotators differ in how they interpret the relationship between texts and hypotheses. 

We investigate both of these possibilities within a \textit{mixed effects modeling} framework \citep{gelman_data_2014}. The core idea is to incorporate annotator-specific parameters into standard NLI models that either (i) merely modify the output of a standard classification/regression head or (ii) modify the parameters of the head itself. These two options correspond to the mixed effects modeling concepts of \textit{random intercepts} and \textit{random slopes}, respectively. For the same reason that such \textit{random effects} can be incorporated into effectively any generalized linear model in a modular way, our components can be be similarly incorporated into any NLI model. We describe how this can be done for a simple RoBERTa-based NLI model. 

We find (i) that models containing only random intercepts outperform both standard models and models containing random slopes when annotators are known; and (ii) that when annotators are not known, performance drops precipitously for both random effects models. Together, these findings suggest that those building NLI datasets should provide annotator information and that those developing NLI systems should incorporate random effects into their models.

\section{Extended Task Definition}
\label{sec:task-definition}

In the standard supervised setting, NLI datasets are (graphs of) functions from text-hypothesis pairs $\thpair{i} \in \Sigma^* \times \Sigma^*$ to inference labels $y_i \in \mathcal{Y}$---where $\mathcal{Y}$ is commonly \{\textit{contradicted}, \textit{neutral}, \textit{entailed}\} or \{\textit{not-entailed}, \textit{entailed}\}, but may also be a finer-grained (e.g. five-point) ordinal scale \citep{zhang-etal-2017-ordinal} or bounded continuous scale \citep{chen-etal-2020-uncertain}. The NLI task is to produce a single label from $\mathcal{Y}$ given a text-hypothesis pair.

We extend this setting by assuming that NLI datasets are (graphs of) functions from text-hypothesis pairs \textit{and} annotator identifiers $a_i \in \annotators$ to inference labels and that the NLI task is to produce a single label given a text-hypothesis pair and an annotator identifier.  A particular model need not make use of the annotator information during training and may similarly ignore it at evaluation time. Though many existing datasets do not provide annotator information, it is trivial for a dataset creator to add (even \textit{post hoc}), and so this extension could feasibly be applied to any existing dataset.

\section{Models}
\label{sec:models}

We assume some encoder that maps from $\thpair{i} \in \Sigma^* \times \Sigma^*$ to $\langle \xb_{T_i}, \xb_{H_i}\rangle \in \reals^M \times \reals^N$ independently of annotator $a_i$, and we focus mainly on the mapping from $\zb_i \equiv \langle \xb_{T_i}, \xb_{H_i}\rangle$ and $a_i$ to $y_i$.

We consider two types of model: one containing only \textit{annotator random intercepts} and another additionally containing \textit{annotator random slopes}. The first assumes that differences among annotators are relatively shallow---e.g. given some true label for a pair (or distribution thereon), annotators have their own specific way of mapping that value to a response---and the second assumes that the differences among annotators are deeper---e.g. annotators differ in how they interpret the relation between texts and hypotheses. This distinction is independent of the labels $\labels$: regardless of whether the labels are discrete or continuous, random effects can be incorporated. In the language of generalized linear mixed models, the \textit{link functions} are the only thing that changes. We consider two label types: three-way ordinal and bounded continuous.

\vspace{-1mm}

\paragraph{Annotator random intercepts} amount to annotator specific bias terms $\bm{\rho}_{a_i}$ on the raw predictions of a classification/regression head. Unlike standard \textit{fixed} bias terms, however, what makes these terms random intercepts is that they are assumed to be distributed according to some prior distribution with unknown parameters. This assumption models the idea that annotators are sampled from some population, and it yields `adaptive regularization' \citep{mcelreath2020statistical}, wherein the biases for annotators who provide few labels will be drawn more toward the central tendency of the prior.

\vspace{-1mm}

\subparagraph{Random intercepts for categorical outputs} can take two forms, depending on whether the model enforces ordinality constraints---as linked logit models do \citep{agresti_categorical_2014}---or not. Since most common categorical NLI models do not enforce ordinality constraints, we do not enforce them here, assuming that the model has some independently tunable function $h_{\bm{\theta}}: \reals^M \times \reals^N \rightarrow \reals^{|\labels|}$ that produces potentials for each label and that: 

\vspace{-7mm}

\[f(y_i\;|\;\zb_i, \bm{\theta}, \bm{\rho}_{a_i}) = \text{softmax}\left(h_{\bm{\theta}}(\zb_i) + \bm{\rho}_{a_i}\right)\]

\vspace{-2mm}

\noindent where $\bm{\rho}_{a_i} \sim \mathcal{N}(\bm{0}, \bm{\Sigma})$ with unknown $\bm{\Sigma}$.

\vspace{-1mm}

\subparagraph{Random intercepts for continuous outputs} are effectively shifting terms on the single value predicted by some independently tunable function $h: \reals^M \times \reals^N \rightarrow \reals$. If the continuous output is furthermore bounded, a squashing function $g$ is necessary. In the bounded case, we assume that the variable---scaled to (0, 1)---is distributed Beta \citep[following][]{sakaguchi-van-durme-2018-efficient} with mean $\mu_i$ and precision $\nu_i = \exp\left(\rho_{a_i1}+\nu_0\right)$. 

\vspace{-7mm}

\begin{align*}
    \mu_i &= g\left(h_{\bm{\theta}}(\zb_i) + \rho_{a_i2}\right)\\
    \alpha_i;\;\beta_i &= \mu_i\nu_i;\;(1-\mu_i)\nu_i\\
    f(y_i\;|\;\zb_i, \bm{\theta}, \bm{\rho}_{a_i}; \nu_0) &= \text{Beta}(y_i\;|\;\alpha_i, \beta_i)
\end{align*}

\vspace{-2mm}

\noindent where $\bm{\rho}_{a_i} \sim \mathcal{N}(\bm{0}, \bm{\Sigma})$ with unknown $\bm{\Sigma}$. This implies that $\nu_i \sim \log\mathcal{N}(\nu_0, \sigma_{11}^2)$ with unknown~$\nu_0$.

The precision parameter $\nu_i$ controls the shape of the Beta: with small $\nu_i$, $a_i$ tends to give responses near 0 and 1 (whichever is closer to $\mu_i$); with large $\nu_i$, $a_i$ tends to give responses near $\mu_i$.

\vspace{-2mm}

\paragraph{Annotator random slopes} amount to annotator-specific classification/regression heads $h_{\bm{\phi}_i}$. We swap these heads into the above equations in place of $h_{\bm{\theta}}$. As for the random intercept parameters, we assume that the annotator-specific parameters $\bm{\phi}_i$, which we refer to as the annotator random slopes, are distributed $\bm{\phi}_i \sim \mathcal{N}(\bm{\theta}, \bm{\Sigma})$ with unknown $\bm{\theta}, \bm{\Sigma}$. One way to think about this model is that $h_{\bm{\theta}}$ produces prototypical interpretation around which annotators' actual interpretations are distributed.

\begin{table}[t]
\small
\centering
    \begin{tabular}{l}
    \toprule
     \multicolumn{1}{c}{\cellcolor{light-gray} \textbf{MegaVeridicality}} \\
     $\blacktriangleright$ \textit{Someone knew that something happened.}\\
       \hspace{1em}\textcolor{ao-eng}{\textit{That thing happened.}}\\
       $\blacktriangleright$ \textit{Someone thought that something happened.}\\
      \hspace{1em}\textcolor{red}{\textit{That thing happened.}}\\
     \midrule      
    \multicolumn{1}{c}{\cellcolor{light-gray} \textbf{MegaNegRaising}} \\
      $\blacktriangleright$ \textit{Someone didn't think that something happened.}\\
       \hspace{1em}\textcolor{ao-eng}{\textit{That person thought that thing didn't happen.}}\\
      $\blacktriangleright$ \textit{Someone didn't know that something happened.}\\
      \hspace{1em}\textcolor{red}{\textit{That person knew that thing didn't happen.}}\\
         \bottomrule
    \end{tabular}
    \vspace{-2mm}
    \caption{NLI sentence pairs from MegaVeridicality and MegNegRaising.
$\blacktriangleright$ indicates the line is a text, and the following line is its corresponding hypothesis. Hypotheses in \textcolor{ao-eng}{green} indicate that the context entails the hypothesis; those in \textcolor{red}{red} indicate that it does not.}
    \label{tab:dataexamples}
    \vspace{-6mm}
\end{table}

\begin{table*}[t]
    \centering
    \begin{tabular}{lcccccccc}
    \toprule
                   & \multicolumn{2}{c}{\textsc{random}} & \multicolumn{2}{c}{\textsc{predicate}} & \multicolumn{2}{c}{\textsc{structure}} & \multicolumn{2}{c}{\textsc{annotator}} \\
    \textbf{Model} & \textit{Acc} & \textit{Corr} & \textit{Acc} & \textit{Corr} & \textit{Acc} & \textit{Corr} & \textit{Acc} & \textit{Corr} \\ 
    \midrule
    Fixed                 & 1.00           & 0.35 & 0.92          & 0.23 & 0.83          & 0.27 & \phantom{-}0.91  & \textbf{0.31}  \\
    Random Intercepts     & 1.15  & \textbf{1.53}  & \textbf{1.13} & \textbf{1.53} & \textbf{1.05} & \textbf{1.53} & \textbf{0.98} & 0.20  \\
    Random Slopes         & \textbf{1.17}           & 1.42 & \textbf{1.13}          & 1.42 & 0.82          & 1.41 & 0.42 & 0.05 \\
    \bottomrule
    \end{tabular}
    \vspace{-2mm}
    \caption{Mean of the rescaled accuracy (categorical data) and rank correlation (bound continuous data) across cross-validation folds for each partitioning method (score$_\text{mod}$ from \S\ref{sec:experiments}). Bolded values are best in column.}
    \label{tab:results}
    \vspace{-5mm}
\end{table*}

\section{Experiments}
\label{sec:experiments}

We compare models both with and without random effects when fit to NLI datasets conforming to the extended setting described in \S\ref{sec:task-definition}. The model without random intercepts (the \textit{fixed model}) simply ignores annotator information---effectively locking $\bm\rho_{a_i}$ to $\bm{0}$ for all annotators $a_i$.

\vspace{-2mm}

\paragraph{Encoder}

All models use pretrained RoBERTa \cite{liu_roberta_2019} as their encoder. We use the basic LM pretrained versions (no NLI fine-tuning).

\vspace{-2mm}

\paragraph{Data}

To our knowledge, the only NLI datasets that both publicly provide annotator identifiers and are large enough to train an NLI system are MegaVeridicality \citep[MV;][]{white_role_2018,white-etal-2018-lexicosyntactic}, which contains three-way categorical annotations aimed at assessing whether different predicates give rise to veridicality inferences in different syntactic structures, and MegaNegRaising \citep[MN;][]{an_lexical_2020}, which contains bounded continuous [0, 1] annotations aimed at assessing whether different predicates give rise to neg(ation)-raising inferences in different syntactic structures. \autoref{tab:dataexamples} shows example pairs from each dataset. Both datasets contain 10 annotations per text-hypothesis pair from 10 different annotators. MV contains 3,938 pairs (39,380 annotations) with 507 distinct annotators, and MN contains 7,936 pairs (79,360 annotations) with 1,108 distinct annotators. In both datasets, each pair is constructed to include a particular main clause predicate and a particular syntactic structure. To test each model's robustness to lexical and structure variability, we use this information to construct folds of the cross-validation (see \textbf{Evaluation}).

\vspace{-2mm}

\paragraph{Classification/Regression Heads}

We consider heads with one hidden affine layer followed by a rectifier. We use a hidden layer size of 128 and the default RoBERTa-base input size of 768.

\vspace{-2mm}

\paragraph{Training}

All models were implemented in PyTorch 1.4.0 and were trained for a maximum of 25 epochs on a single Nvidia GeForce GTX 1080 Ti GPU, with early stopping upon a change in average per-epoch loss of less than 0.01. We use Adam optimization (lr=0.01, $\beta_1$=0.9, $\beta_2$=0.999, $\epsilon$=10$^{-7}$) and a batch size of 128. All code is \href{https://github.com/wgantt/nli-mixed-models}{publicly available}.

\vspace{-2mm}

\paragraph{Loss}

We use the negative log-likelihood of the observed values under the model as the loss.

\vspace{-2mm}

\paragraph{Evaluation} 

We evaluate all of our models using 5-fold cross-validation. We consider four partitioning methods: (i) \textsc{random}: completely random partitioning; (ii) \textsc{predicate}: partitioning by the main clause predicate found in the text (a particular main clause predicate occurs in one and only one partition); (iii) \textsc{structure}: partitioning by the syntactic structure found in the text (a particular structure occurs in one and only one partition); and (iv) \textsc{annotator}: a particular annotator occurs in one and only one partition. For the first three methods, we ensure that each annotator occurs in every partition, so that random intercepts and random slopes for that annotator can be estimated. For the \textsc{annotator} method, where we do not have an estimate for the random effects of annotators in the held-out data, we use the mean of the prior.\footnote{We additionally experimented with marginalizing over the random effects, but the results did not differ.} 

We report mean accuracy on held-out folds for the categorical data (MV); and following \citet{chen-etal-2020-uncertain}, we report mean rank correlation on held-out folds for the bounded continuous data (MN). To make these metrics comparable, we report them relative to the performance of both a baseline model and the best possible fixed model.

\vspace{-6mm}

\[\text{score}_\text{mod} = \frac{\text{raw-score}_\text{mod} - \text{raw-score}_\text{base}}{\text{raw-score}_\text{best} - \text{raw-score}_\text{base}}\]

\vspace{-2mm}

\noindent For the categorical data, the baseline model predicts the majority class across all pairs, and the best possible fixed model predicts the majority class across annotators for each pair. Similarly, for the bounded continuous data, the baseline model predicts the mean response across all pairs, and the best possible fixed model predicts the mean response across annotators for each pair.\footnote{Rank correlation is technically undefined when one of the variables is constant. For the purposes of computing $\text{score}_{\text{mod}}$ for the bounded continuous data, we treat $\text{raw-score}_{\text{base}}$ as 0.}

These relative scores are 0 when the model does not outperform the baseline and 1 when the model performs as well as the best possible fixed model. It is possible for a random effects model to obtain a score of greater than 1 by leveraging annotator information or less than 0 if it overfits the data.

\section{Results}
\label{sec:results}

\autoref{tab:results} shows the results. The random intercepts models reliably outperform the fixed models in all cross-validation settings except \textsc{annotator} in Bonferroni-corrected Wilcoxon rank-sum tests ($p$s$<$0.05). Indeed, they tend to reliably outperform even the best possible fixed model, having rescaled scores above 1. The random slopes models, while in many cases comparable to the random intercepts models, confer no additional benefit over them. In the one instance in which the random slopes model performs best (the random partition for categorical data), the advantage relative to the random intercepts model is not statistically significant.

Consistent with \citeauthor{pavlick-kwiatkowski-2019-inherent}'s findings, these results suggest that variability in annotators' responding behavior is substantial; otherwise, it would not be possible for the random effects models to outperform the best possible fixed model, and we would not expect the observed drops in performance when annotator information is removed. But this variability is likely relatively shallow: if these differences were due to deeper differences in annotators' interpretation of the pair, we would expect this to manifest in better performance by the random slopes models, as the latter subsumes the random intercepts model and can leverage the additional power of annotator-specific classification or regression heads. Of course, it remains a live possibility that the encoder we used does not extract features that are linearly related to the relevant interpretive variability, and so further investigation of random slopes models with different encoders may be warranted \citep[see][]{geva-etal-2019-modeling}.

Contrasting the results on ordinal and bounded continuous data, the fixed model tends to perform better on ordinal data than on bounded continuous data. A similar trend is not seen for the random effects models. Indeed, the random intercepts model performs substantially better on the bounded continuous data under all settings except for \textsc{annotator}. These results could be due to the link function we used for the bounded continuous data: the fixed model consistently learned small values for the precision parameter $\nu_0$, resulting in sparse (bimodal) beta distributions. But the fact that the random intercepts model reliably outperforms the best possible fixed model implies that any tweaks to the link function would not bring the fixed model up to the level of the random intercepts model. 

\section{Analysis}
\label{sec:analysis}

\begin{figure}
    \centering
    \vspace{-7mm}
    \includegraphics[width=\columnwidth]{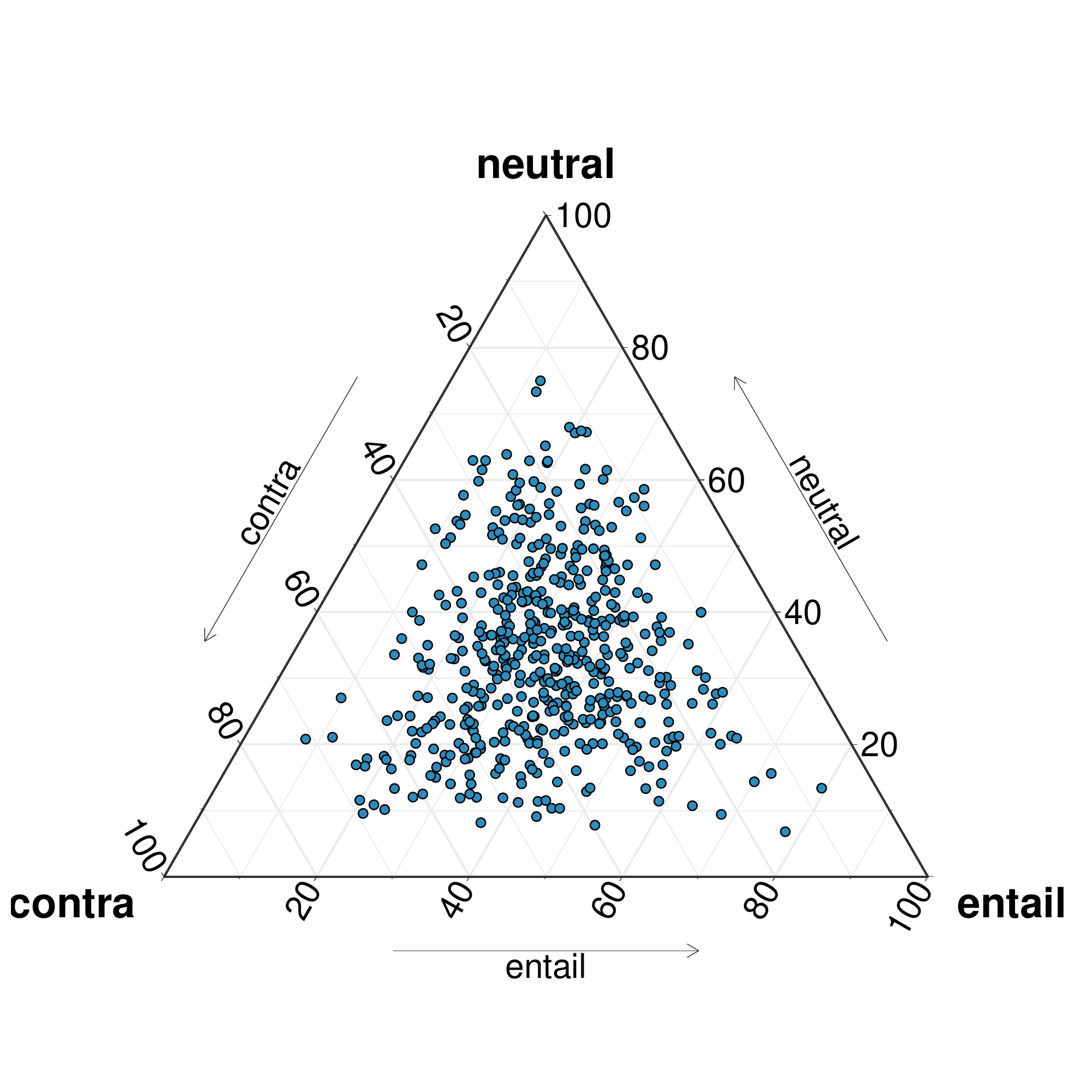}
    \vspace{-15mm}
    \caption{Distribution of biases across categorical annotators when fixed effect potentials set to $\mathbf{0}$.}
    \label{fig:categorical-bias}
    \vspace{-5mm}
\end{figure}

To understand how annotator biases tend to pattern with ordinal and bounded continuous scales, we investigate the mean $\bm\rho_a$ for each annotator $a$ in the random intercepts models across folds under the \textsc{random} partition method. \autoref{fig:categorical-bias} plots the distribution of biases across categorical annotators when the fixed effect potentials---$h_{\bm{\theta}}(\zb_i)$ in the equations in \S\ref{sec:models}---are set to $\mathbf{0}$: softmax($\bm\rho_a$). This distribution can be thought of as an indicator of how an annotator would respond in the absence of any correct answer. We see the most variability in terms of annotators' biases for or against neutral: the interquartile range for \textit{neutral} biases is [0.23, 0.42] compared to [0.24, 0.41] for \textit{contradiction} and [0.25, 0.40] for \textit{entailment}. Interestingly, these biases do not reflect the fact that the scale is ordinal: if they did, we would expect more positive correlations between adjacent values; but \textit{neutral} biases are more strongly rank anticorrelated with \textit{contradiction} ($r$ = $-$0.57) and \textit{entailment} ($r$ = $-$0.48) than \textit{contradiction} is with \textit{entailment} ($r$ = $-$0.35). This finding suggests that three-value ``ordinal'' NLI scales are better thought of as nominal.   

\begin{figure}
    \centering
    \includegraphics[width=.92\columnwidth]{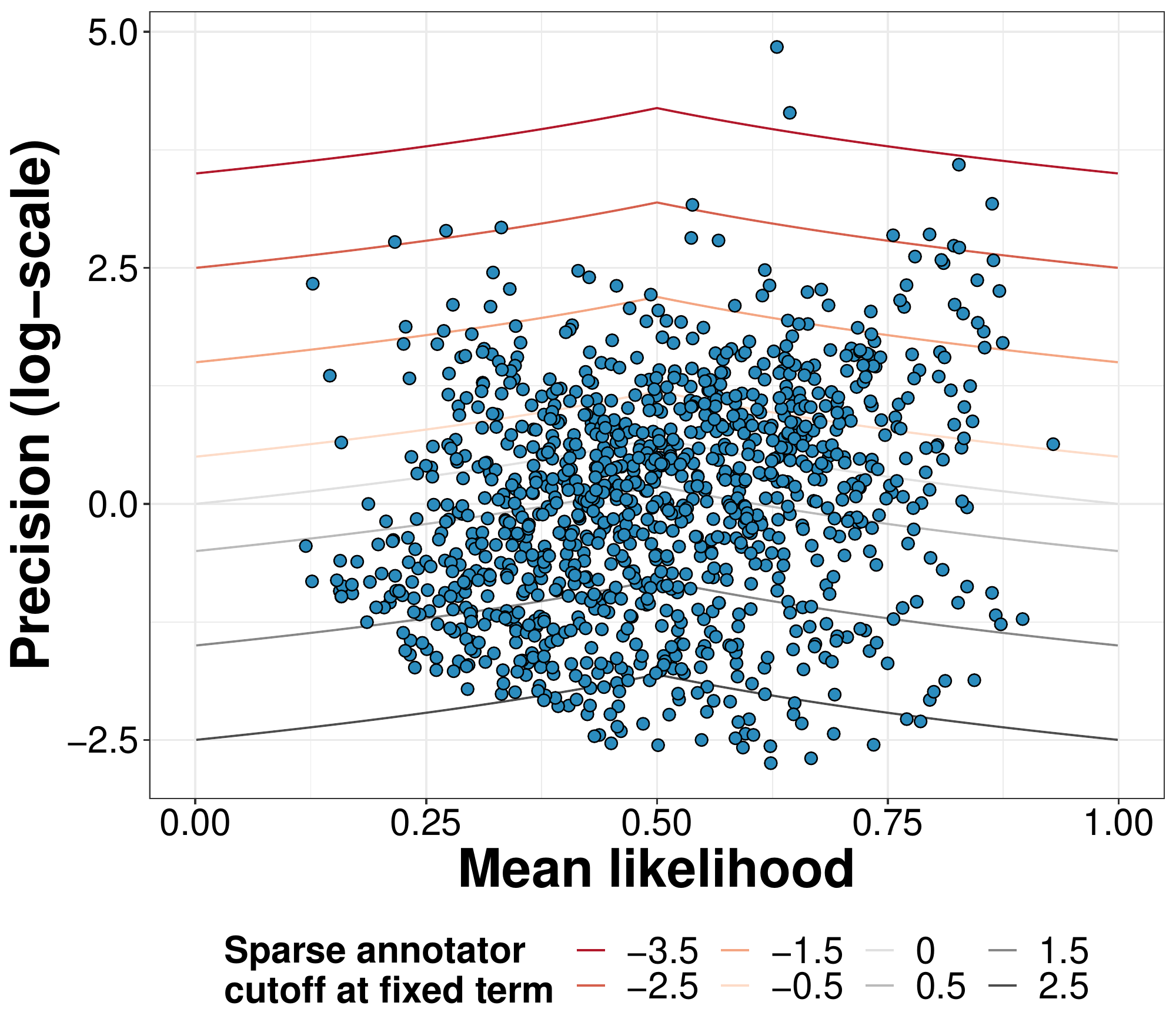}
    \vspace{-2mm}
    \caption{Distribution of biases across bounded continuous annotators when fixed effect potentials set to $\mathbf{0}$.}
    \label{fig:unit-bias}
    \vspace{-5mm}
\end{figure}

\autoref{fig:unit-bias} plots the analogous distribution for the bounded continuous annotators, with the $y$-axis showing $\rho_{a1}$ and the $x$-axis showing logit$^{-1}(\rho_{a2})$. The lines behind the points show, for particular values of $h_{\bm{\theta}}(\zb_i)$, the $\rho_{a1}$ at which the distribution for a particular annotator becomes sparse---i.e. where $\alpha, \beta < 1$---heavily favoring responses very near zero or one, rather than the mean. We see a weak rank correlation ($r$ = 0.24, $p <$ 0.05) between precision and annotators' biases to give responses nearer to one, suggesting that one-biased annotators tend to give less sparse responses. This correlation might, in part, explain the poor performance of the bounded continuous models in the \textsc{annotator} cross-validation setting.

\section{Related Work}
\label{sec:related-work}

The models developed here are closely related to models from Item Response Theory (IRT). IRT has been used to assess annotator quality \citep{hovy-etal-2013-learning, hovy-etal-2014-experiments, rehbein-ruppenhofer-2017-detecting, paun-etal-2018-comparing, paun-etal-2018-probabilistic,zhang-etal-2019-evidence, Felt:Ringger:Seppi:Boyd-Graber-2018} and various properties of an item
\citep{passonneau-carpenter-2014-benefits,sakaguchi-van-durme-2018-efficient,card-smith-2018-importance}, including difficulty
\citep{lalor-etal-2016-building,lalor-etal-2018-understanding,lalor-etal-2019-learning}. Other non-IRT-based work attempts to measure the relationship between annotator disagreement and item difficulty \citep{plank-etal-2014-linguistically,kalouli-etal-2019-explaining}.

Other recent work focuses on incorporating annotator information in modeling annotator-generated text. \citet{geva-etal-2019-modeling} find that concatenating annotator IDs as input features to a BERT-based text generation model yields improved performance on several datasets. Although we reach similar conclusions about the importance of annotator information in this work, our approach differs in at least one critical respect: by explicitly distinguishing linguistic input from annotator information, our model cleanly separates the linguistic representations from representations of the annotators interpreting or producing those representations. This clean separation is of potential benefit not only to those interested in using NLI models (or deep learning architectures more generally) in an experimental (psycho)linguistics setting, where distinguishing the two sorts of representations can be crucial, but also to those interested in possibly quite substantial reductions in model size.

\section{Conclusion}
\label{sec:conclusion}

We find (i) that models containing only random intercepts outperform standard models when annotators are known, and (ii) that models that further contain random slopes do not yield any additional benefit. These results indicate that, though differences among NLI annotators' response behavior are important to model, these differences may not be particularly deep, limited to the ways in which annotators use the response scale, but not relating to deeper interpretive differences.

\section*{Acknowledgments}

This research was supported by the University of Rochester, DARPA AIDA, DARPA KAIROS, IARPA BETTER, and NSF-BCS (1748969).  The U.S. Government is authorized to reproduce and distribute reprints for Governmental purposes. The views and conclusions contained in this publication are those of the authors and should not be interpreted as representing official policies or endorsements of DARPA or the U.S. Government.

\bibliography{acl2020.bbl}
\bibliographystyle{acl_natbib}

\end{document}